\documentclass[10pt,twocolumn,letterpaper]{article}

 \usepackage[pagenumbers]{cvpr} %

\definecolor{cvprblue}{rgb}{0.21,0.49,0.74}
\usepackage[pagebackref,breaklinks,colorlinks,allcolors=cvprblue]{hyperref}

\usepackage{booktabs}
\usepackage{multirow}
\usepackage{multicol}
\usepackage{color, colortbl}

\def\paperID{7035} %
\def\confName{CVPR}
\def\confYear{2025}

\title{GaussianAD: Gaussian-Centric End-to-End Autonomous Driving}

\author{
Wenzhao Zheng$^{1,}$\footnotemark[1] $^{,}$\footnotemark[2] \quad 
Junjie Wu$^{2,}$\footnotemark[1] \quad 
Yao Zheng$^{2,}$\footnotemark[1] \quad 
Sicheng Zuo$^{1}$ \quad 
Zixun Xie$^{2,3}$ \quad  \\
Longchao Yang$^{2}$ \quad
Yong Pan$^2$ \quad 
Zhihui Hao$^2$ \quad
Peng Jia$^2$ \quad
Xianpeng Lang$^2$ \quad
Shanghang Zhang$^{3,}$\footnotemark[3] \\
$^1$Tsinghua University \quad 
$^2$Li Auto \quad
$^3$Peking University  \\
\texttt{wenzhao.zheng@outlook.com; shanghang@pku.edu.cn} \\
Project Page: \url{https://wzzheng.net/GaussianAD} \\
Large Driving Models: \url{https:/github.com/wzzheng/LDM}
}

\begin{document}

\twocolumn[{%
\renewcommand\twocolumn[1][]{#1}%
\vspace{-8mm}
\maketitle
\vspace{-5mm}
}]

\renewcommand{\thefootnote}{\fnsymbol{footnote}}
\footnotetext[1]{Equal contributions. $\dagger$Project leader. $\ddagger$Corresponding author.}
\renewcommand{\thefootnote}{\arabic{footnote}}

\begin{abstract}
Vision-based autonomous driving shows great potential due to its satisfactory performance and low costs.
Most existing methods adopt dense representations (e.g., bird's eye view) or sparse representations (e.g., instance boxes) for decision-making, which suffer from the trade-off between comprehensiveness and efficiency. 
This paper explores a Gaussian-centric end-to-end autonomous driving (GaussianAD) framework and exploits 3D semantic Gaussians to extensively yet sparsely describe the scene.
We initialize the scene with uniform 3D Gaussians and use surrounding-view images to progressively refine them to obtain the 3D Gaussian scene representation.
We then use sparse convolutions to efficiently perform 3D perception (e.g., 3D detection, semantic map construction).
We predict 3D flows for the Gaussians with dynamic semantics and plan the ego trajectory accordingly with an objective of future scene forecasting. 
Our GaussianAD can be trained in an end-to-end manner with optional perception labels when available. 
Extensive experiments on the widely used nuScenes dataset verify the effectiveness of our end-to-end GaussianAD on various tasks including motion planning, 3D occupancy prediction, and 4D occupancy forecasting.
Code: \url{https://github.com/wzzheng/GaussianAD}.

\end{abstract}
     
\section{Introduction}
\label{sec:intro}

Vision-based autonomous driving emerges as a promising direction due to its resemblance with human driving and economic sensor configuration~\cite{caddn,bevdepth,lss,bevdet,bevformer}.
Despite the lack of depth inputs, vision-based methods exploit deep networks to infer structural information from RGB cameras and demonstrate strong performance in various tasks, such as 3D object detection~\cite{bevdepth,bevdet,bevformer}, HD map construction~\cite{liao2022maptr,liu2022vectormapnet,li2022hdmapnet,beverse}, and 3D occupancy prediction~\cite{tpvformer,surroundocc,tian2023occ3d,tong2023scene,openoccupancy,pointocc,gaussianformer-2}.

\begin{figure}[t]
\centering
\includegraphics[width=0.475\textwidth]{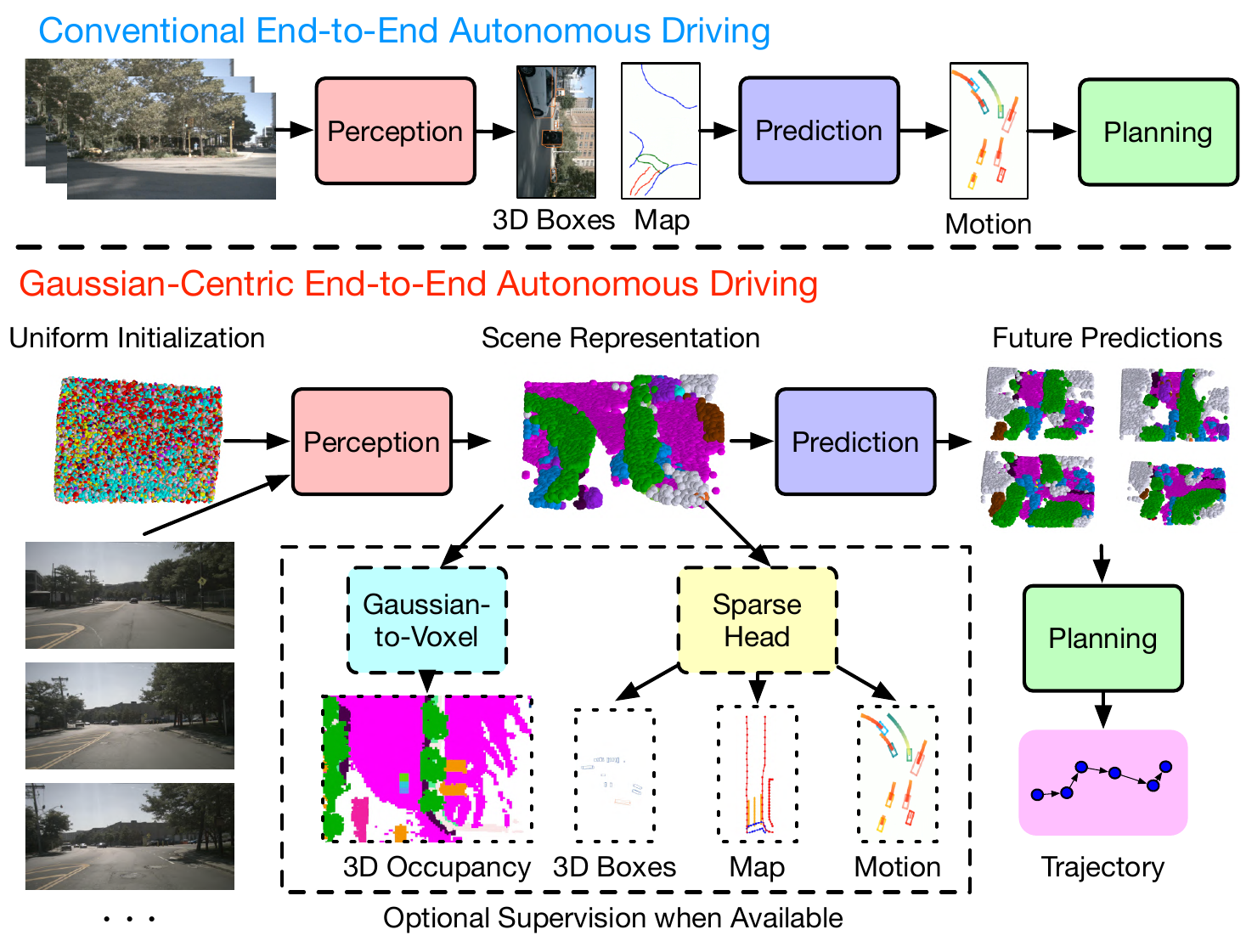}
\vspace{-9mm}
\caption{\textbf{Comparisons of different pipelines for autonomous driving.}
Conventional end-to-end autonomous driving methods usually obtain refined scene descriptions (e.g., 3D boxes, maps) as the interface for prediction and planning, which may omit certain critical information.
Differently, the proposed GuassianAD employs sparse yet comprehensive 3D Gaussians to pass information through the pipeline to efficiently preserve more details.
We can optionally impose dense or sparse supervision to instruct the learning of scene representations.
Our pipeline can adapt to various data with different available annotations.
}
\label{fig:teaser}
\vspace{-6mm}
\end{figure}

Recent autonomous driving research is undergoing a shift from the modular~\cite{tpvformer,beverse,bevformer} to the end-to-end paradigm, which aims to plan the future trajectory directly from image inputs~\cite{hu2022stp3,uniad,vad,ye2023fusionad,genad}.
The key advantage of the end-to-end pipeline is less information loss from the inputs to the outputs, making it important to design the intermediate 3D scene representation of 2D images.
Conventional methods compress the 3D scene in the height dimension to obtain the bird's eye view (BEV) representation~\cite{uniad, ye2023fusionad}.
Recent methods explore sparse queries (e.g. instance boxes, map elements) to describe the surrounding scene~\cite{sparsead,sparsedrive}.
Despite their efficiency, they cannot capture the fine-grained structure of the 3D environment, providing less knowledge to the decision-making process.
Furthermore, some methods employ tri-perspective view~\cite{tpvformer,s2tpvformer,pointocc} or voxels~\cite{surroundocc,tong2023scene,occworld} to represent scenes as 3D occupancy to capture more comprehensive details.
However, the dense modeling leads to large computation overhead and thus fewer resources to reason about decision-making. 
This raises a natural question: 
\emph{can we design a comprehensive yet sparse intermediate representation to pass information through the end-to-end model?}

This paper proposes a Gaussian-centric autonomous driving (GaussianAD) framework as a positive answer, as shown in Figure~\ref{fig:teaser}.
We employ a sparse set of 3D semantic Gaussians~\cite{gaussianformer} from 2D images as the scene representations.
Despite the sparsity, it benefits from fine-grained modeling resulting from the universal approximation of Gaussian mixtures and explicit 3D structure facilitating various downstream tasks.
We further explore perception, prediction, and planning from the 3D Gaussian representation.
For perception, we treat 3D Gaussians as semantic point clouds and employ sparse convolutions and sparse prediction heads to efficiently process the 3D scene. 
We propose 3D Gaussian flow to comprehensively and explicitly model the scene evolution, where we predict a future displacement for each Gaussian.
We then integrate all available information to plan the ego trajectory accordingly.
Due to the explicitness of 3D Gaussian representation, we can straightforwardly compute the forecasted future scenes observed by the ego car using affine transformations. 
We compare the forecasted scenes with ground-truth scene observations as explicit supervision for both prediction and planning. 
To the best of our knowledge, our GaussianAD is the first to explore the explicitly sparse point-based architecture for vision-centric end-to-end autonomous driving. 
We conduct extensive experiments on the nuScenes~\cite{nuscenes} dataset to evaluate the effectiveness of the proposed Gaussian-centric framework. 
Experimental results demonstrate that our GaussianAD achieves state-of-the-art results on end-to-end motion planning with high efficiency.

\section{Related Work}
\label{sec:formatting}

\textbf{Perception for Autonomous Driving.}
Accurately perceiving the surrounding environment from sensor inputs is the fundamental step for autonomous driving.
As the two main conventional perception tasks, 3D object detection aims to obtain the 3D position, pose, and category of each agent in the surrounding scene~\cite{caddn,bevdepth,lss,bevdet,beverse,bevformer}, which are important for trajectory prediction and planning.
Semantic map reconstruction aims to recover the static map elements in the bird's eye view (BEV) to provide additional information for further inference~\cite{liao2022maptr,liu2022vectormapnet,li2022hdmapnet,beverse}.
Both tasks can be efficiently performed in the BEV space, yet they cannot describe the fine-grained 3D structure of the surrounding scene and arbitrary-shape objects~\cite{tpvformer,surroundocc}.
This motivates recent methods to explore other 3D representations like voxel and tri-perspective view (TPV)~\cite{tpvformer} to perform the 3D occupancy prediction task~\cite{surroundocc,tian2023occ3d,tong2023scene,openoccupancy,pointocc}.
3D occupancy provides more comprehensive descriptions of the surrounding scene including both dynamic and static elements, which can be efficiently learned from sparse LiDAR~\cite{tpvformer} or video sequences~\cite{cao2023scenerf}.
Gaussianformer~\cite{gaussianformer} proposed to use 3D semantic Gaussians to represent the scene for 3D occupancy sparsely.
However, it is still not clear whether the 3D Gaussian representation can be used for general autonomous driving.

\textbf{Prediction for Autonomous Driving.}
Predicting the scene evolution is also vital to the safety of autonomous driving vehicles.
Most existing methods focus on predicting the movement of traffic agents given their past positions and semantic map information~\cite{phan2020covernet,beverse,liu2021mmtrans,hu2021fiery,gu2022vip3d, jiang2022pip,gpd-1}.
Early methods projected agent and semantic map information onto BEV images and employed 2D image backbones to process them to infer future agent motions~\cite{chai2019multipath, phan2020covernet}.
Subsequent methods adopted a more efficient tokenized representation of dynamic agents and used graph neural networks~\cite{liang2020lanegcn} or transformers~\cite{vaswani2017attention, liu2021mmtrans, ngiam2021scenetrans} to aggregate information.
Recent works began to explore motion prediction directly from sensor inputs in an end-to-end manner~\cite{hu2021fiery, beverse, gu2022vip3d, jiang2022pip,vad}.
They usually first perform BEV perception to extract relevant information (e.g., 3D agent boxes, semantic maps, tracklets) and then exploit them to infer future trajectories.
Different from existing methods which only model dynamic object motions, we propose Gaussian flows to predict the surrounding scene evolutions including both dynamic and static elements.

\begin{figure*}[t]
\centering
\includegraphics[width=0.95\textwidth]{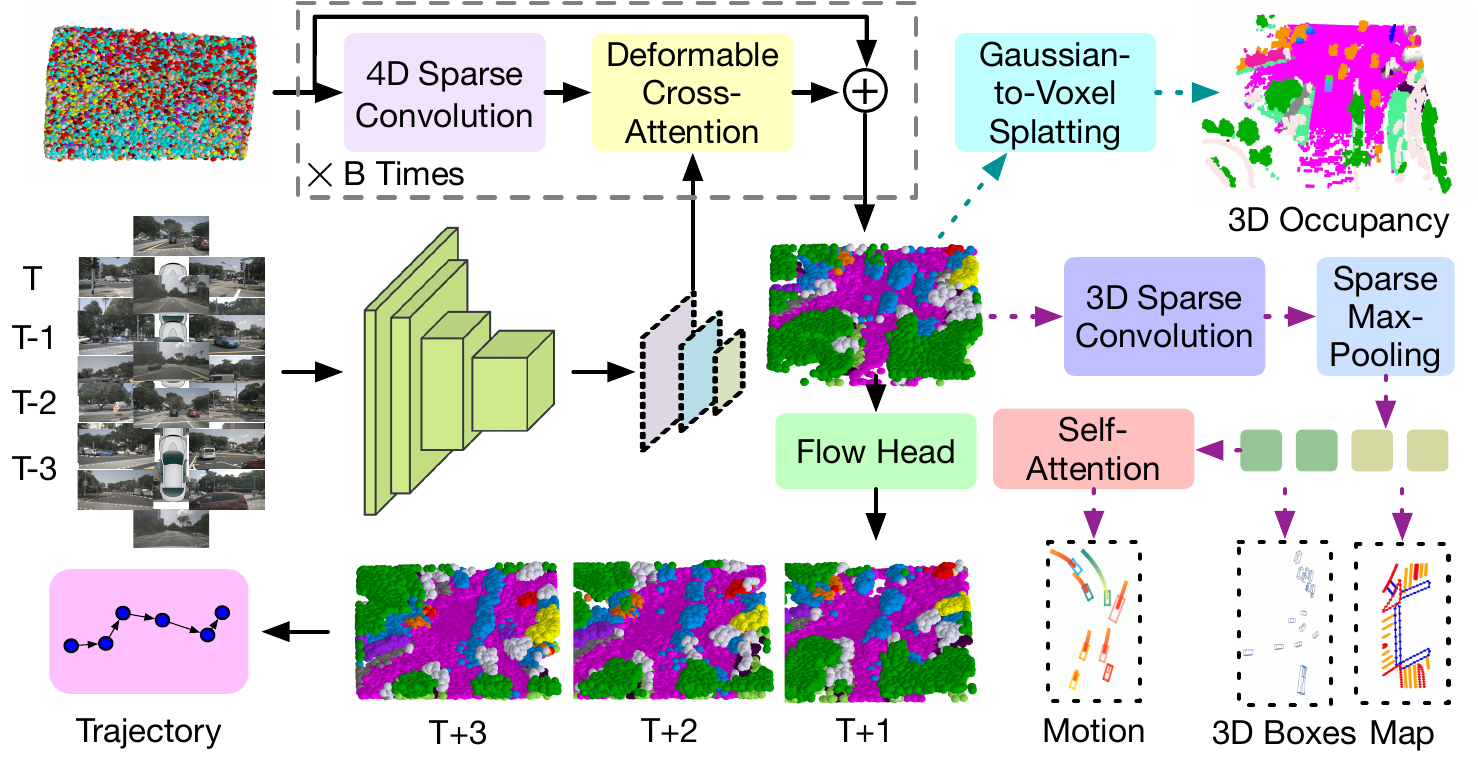}
\vspace{-5mm}
\caption{\textbf{Overview of the proposed GaussianAD framework.}
We initialize the sequence of 3D scenes with uniform Gaussians and employ 4D sparse convolutions to enable interactions between Gaussians.
We then extract multi-scale features from surrounding-view multi-frame image observations and use deformable cross-attention to incorporate them into the 3D Gaussians. 
Having obtained the temporal 3D Gaussians as the scene representation, we can optionally employ Gaussian-to-voxel splatting~\cite{gaussianformer} for dense tasks (e.g., 3D semantic occupancy) or use sparse convolutions and max-pooling~\cite{voxelnext} for sparse tasks (e.g., 3D object detection, HD map construction, motion prediction).
We use a flow head to predict a 3D flow for each Gaussian and aggregate them for trajectory planning.
}
\label{fig:framework}
\vspace{-6mm}
\end{figure*}

\textbf{Planning for Autonomous Driving.} 
Planning is the essential component of autonomous driving systems, which can be categorized into rule-based~\cite{treiber2000idm,bouchard2022rule,Dauner2023CORL} and learning-based~\cite{scheel2022urban, cheng2022mpnp, pini2023safepathnet} methods.
While the traditional rule-based methods can achieve satisfactory results with high interpretability~\cite{Dauner2023CORL}, learning-based methods have received increasing attention in recent years due to their great potential to scale up to large-scale training data~\cite{ bronstein2022hierarchical, couto2023hierarchical,liu2022improved, huang2022efficient,zhou2021exploring,wen2020fighting}.
As simple yet effective learning-based solutions, imitation-based planners have been the preferred choices for end-to-end methods~\cite{zhou2021exploring, vitelli2022safetynet, huang2023gameformer, huang2023dipp, guo2023ccil,cheng2023rethinking}.
As early attempts, LBC~\cite{chen2020lbc} and CILRS~\cite{codevilla2019cilrs} employed convolutional neural networks (CNNs) to learn from expert driving data.
The following methods incorporated more data~\cite{ye2023fusionad} or extracted more intermediate features~\cite{hu2022stp3,uniad,vad,genad} to provide more information for the planner, which achieved remarkable performance.
Still, most existing end-to-end autonomous driving methods adopt high-level scene descriptions (e.g., 3D boxes, maps) for downstream prediction and planning and may omit certain critical information.
This paper proposes a Gaussian-centric autonomous driving pipeline and uses 3D Gaussians as sparse yet comprehensive information carrier.

\section{Proposed Approach}

\subsection{3D Scene Representation Matters for Driving} \label{sec1}
Autonomous driving aims to produce safe and consistent control signals (e.g., accelerator, brake, steer) given a series of scene observations $\{ \mathbf{o} \}$.
While the scene observations $\{ \mathbf{o} \}$ can be obtained from multiple sensors such as cameras and LiDAR, we mainly target vision-based autonomous driving from surrounding cameras due to its high information density and low sensor costs~\cite{hu2022stp3,uniad,vad,ye2023fusionad,genad,tong2023scene}.

Assuming a good-performing controller, most autonomous driving models mainly focus on learning the mapping $f$ from the current and history observations $\{ \mathbf{o} \}$ to the future ego trajectories $\{ \mathbf{w} \}$:
\begin{equation}\label{eqn: AD_model}
\begin{aligned}
   \{ \mathbf{o}^{T-H}, \cdots, \mathbf{o}^{T} \} \xrightarrow{f} \{ \mathbf{w}^{T+1}, \cdots, \mathbf{w}^{T+F} \},
\end{aligned}
\end{equation}
where $T$ denotes the current time stamp, $H$ is the number of history frames, and $F$ is the number of predicted future frames.
Each waypoint $\mathbf{w}=\{x, y, \psi \}$ is determined by the 2D position $\{x, y\}$ and the yaw angle $\psi$ (i.e., the advancing direction of the ego vehicle) in the bird's eye view (BEV).

Conventional autonomous driving methods decompose $f$ into perception, prediction, and planning modules and train them separately before connecting~\cite{bevformer, tpvformer, chai2019multipath, phan2020covernet, Dauner2023CORL}:
\begin{eqnarray} \label{eqn: modular}
   &&\text{Perception:} \  \{ \mathbf{o}^{T-H}, \cdots, \mathbf{o}^{T} \} \xrightarrow{} \mathbf{d}^T, \nonumber \\
   &&\text{Prediction:} \  \mathbf{d}^T \xrightarrow{} \{ \mathbf{d}^{T+1}, \cdots, \mathbf{d}^{T+F} \}, \\
   &&\text{Planning:} \  \{ \mathbf{d}^{T+1}, \cdots, \mathbf{d}^{T+F} \} \xrightarrow{} \{ \mathbf{w}^{T+1}, \cdots, \mathbf{w}^{T+F} \} \nonumber,
\end{eqnarray}
where $\mathbf{d}$ is the scene description such as instance bounding boxes of other agents or map elements of the surroundings.
The scene description $\mathbf{d}$ usually only provides a partial representation of the scene, resulting in information loss.

The separate training of these modules further aggravates this issue as different tasks focus on extracting different information.
The incomprehensive information provided to the planning module might bias the decision-making process of the autonomous driving model.
This motivates the shift from the modular framework to the end-to-end framework~\cite{uniad,genad,sparsead}, which differentiably bridges and jointly learns the perception, prediction, and planning modules:
\begin{equation}\label{eqn: e2e}
\begin{aligned}
  & \{ \mathbf{o}^{T-H}, \cdots, \mathbf{o}^{T} \} \xrightarrow{} \mathbf{r}^T \xrightarrow{} \mathbf{r}^T, \mathbf{d}^T \xrightarrow{} \\
  & \mathbf{r}^T, \{ \mathbf{d}^{T+1}, \cdots, \mathbf{d}^{T+F} \} \xrightarrow{} \{ \mathbf{w}^{T+1}, \cdots, \mathbf{w}^{T+F} \},
   \end{aligned}
\end{equation}
where $\mathbf{r}$ is the scene representation.
$\mathbf{r}$ is usually composed of a set of continuous features and provides a more comprehensive representation of the 3D scenes than $\mathbf{d}$.

The scene representation $\mathbf{r}$ conveys information throughout the model, making the choice of $\mathbf{r}$ critical to the performance of the end-to-end system.
As autonomous driving needs to make decisions in the 3D space, the scene representation should be 3D-structured and contain 3D structural information inferred from the input images.
On the other hand, 3D space is usually sparse, resulting in a tradeoff between comprehensiveness and efficiency when designing $\mathbf{r}$.

For comprehensiveness, the conventional bird's eye view (BEV) representation~\cite{bevdepth,bevformer,beverse} uses dense grid features in the map view and compresses the height dimension to reduce redundancy.
Subsequent methods further explore more dense representations such as voxels~\cite{surroundocc} or tri-perspective view (TPV)~\cite{tpvformer} to capture more detailed and fine-grained 3D information.
For efficiency, recent methods~\cite{sparsead,sparsedrive} adopt sparse queries and focus on modeling instance boxes and map elements, which are the most important factors for decision-making.
Still, the discarded information can still be important (e.g., irregular obstacles, traffic lights, human poses) and is contradictory to the philosophy of end-to-end autonomous driving (i.e., comprehensive information flow). 
This paper explores the 3D Gaussians as a comprehensive yet sparse scene representation and proposes a fully sparse framework for end-to-end perception, prediction, and planning, as shown in Figure~\ref{fig:framework}.

\subsection{Gaussian-Centric Autonomous Driving}

\textbf{3D Gaussian Representation.} Existing methods typically build a dense 3D feature to represent the surrounding environment and processes every 3D voxel with equal storage and computation resources, which often leads to intractable overhead because of unreasonable resource allocation. At the same time, this dense 3D voxel representation cannot distinguish objects of different scales. Unlike these methods, we follow  GaussianFormer \cite{gaussianformer} which represents an autonomous driving scene with a number of  sparse 3D semantic Gaussians. Each Gaussian instantiates a semantic Gaussian distribution characterized by mean, covariance, and semantic logits. This sparse explicit feature representation is more beneficial for downstream tasks. 

\textbf{Gaussians From Images.}
We first represent 3D Gaussians and their high-dimensional queries as learnable vectors. We then employ a Gaussian encoder to iteratively enhance these representations. Each Gaussian Encoder block is composed of three modules: a self-encoding module facilitating interactions between Gaussians, an image cross-attention module for aggregating visual information, and a refinement module to fine-tune Gaussian properties. 
Different from GaussianFormer~\cite{gaussianformer}, we utilize a temporal encoder consisting of 4D sparse convolutions to integrate Gaussian features from the previous frame with the corresponding features in the current frame.

\textbf{Sparse 3D Detection from Gaussians.} As 3D Gaussian representation is a sparse scene representation, we follow VoxelNeXt~\cite{voxelnext} which predicts 3D objects directly based on sparse voxel features. Specially, we conduct a 3D sparse CNN network $\mathbf{V}$ to encode 3D Gaussian representation $\mathbf{r}$. Following GenAD~\cite{genad}, we decode 3D objects $\mathbf{a}$ with a set of agent tokens $\mathbf{D}$ on $\mathbf{V}(\mathbf{r})$:
\begin{equation}\label{eqn: Sparse_3D}
\begin{aligned}
   \mathbf{a}=f_{a}(\mathbf{D}, \mathbf{V}(\mathbf{r})),
\end{aligned}
\end{equation} 
where $f_{a}$ represents a combine of global cross-attention mechanism to learn 3D objects tokens and a 3D object decoder head $\mathbf{d_{a}}$ on learned 3D objects tokens.

\textbf{Sparse Map Construction from Gaussians.} Similar to the representation of 3D detection from Gaussian, we adopt a set of map tokens $\mathbf{M}$ to represent semantic maps. 
We focus on three categories of map elements(i.e., lane divider, road boundary, and pedestrian crossing).
\begin{equation}\label{eqn: Sparse_Map}
\begin{aligned}
   \mathbf{m}=f_{m}(\mathbf{M}, \mathbf{V}(\mathbf{r})),
\end{aligned}
\end{equation}
where $f_{m}$ represents a combination of a global cross-attention mechanism to learn map tokens and a semantic map elements decoder head $\mathbf{d_{m}}$ on learned map tokens.

\textbf{Motion Prediction.} Motion prediction module assists ego trajectories planning by forecasting the future trajectories of other traffic participants ~\cite{uniad}. We obtain motion tokens $\mathbf{M_{o}}$ by make agent tokens $\mathbf{D}$ interact with map tokens $\mathbf{M}$ through cross-attention layers $CA$:
\begin{equation}\label{eqn: motion_pred}
\begin{aligned}
   \mathbf{M_{o}}=CA(\mathbf{D}, \mathbf{M}).
\end{aligned}
\end{equation}
A motion decoder $\mathbf{d_{mo}}$ can be applied on motion tokens $\mathbf{M_{o}}$, meanwhile the learned motion tokens $\mathbf{M_{o}}$ are fed to ego trajectory planning head.

\textbf{Gaussian Flow for Scene Prediction.} Furthermore, it shows that the scene prediction of the intermediate representation $\mathbf{r}$ plays a significant role in end-to-end autonomous driving~\cite{occworld}. We predict the future Gaussian representation as Gaussian flow $\mathbf{{r}^{T+N}}$ from current Gaussian representation $\mathbf{{r}^{T}}$ and the predicted ego trajectories $\mathbf{w}^{T+N}$:
\begin{equation}\label{eqn: gauss_flow}
\begin{aligned}
   \mathbf{r}^{T+N}=f_{r}(\mathbf{r}^{T}, \mathbf{w}^{T+N}).
\end{aligned}
\end{equation}
We then feed the predicted future Gaussian representation $\mathbf{{r}^{T+N}}$ to an occupancy decoder $\mathbf{d}_{occ}$ ~\cite{gaussianformer} to predict future occupancy. The supervision of future occupancy on the intermediate Gaussian representation guarantees the scene forecasting ability which finally improves the performance of ego trajectory prediction.

\subsection{End-to-End GaussianAD Framework}

This subsection presents the overall end-to-end framework of our proposed GaussianAD.
We first initialize the scenes with a set of uniformly distributed 3D Gaussians $\mathbf{G}_0$ and then progressively refine them by incorporating information from the surrounding-view images $\mathbf{o}$ to obtain the Gaussian scene representation $\mathbf{r}$.
We can then optionally extract various scene descriptions $\mathbf{d}$ from $\mathbf{r}$ as auxiliary tasks if the corresponding annotations are available. 
Concretely, we employ Gaussian-to-voxel splatting~\cite{gaussianformer} to obtain dense voxel features for dense descriptions (e.g., 3D occupancy prediction) and fully sparse convolutions~\cite{voxelnext} to obtain sparse queries for sparse descriptions (e.g., 3D bounding boxes, map elements). 
The use of auxiliary perception supervisions introduces additional constraints and prior knowledge on the scene representation $\mathbf{r}$ to guide its learning process.
Still, we predict future evolutions directly on the 3D Gaussians $\mathbf{r}$ to reduce information loss and plan the ego trajectory $\{ \mathbf{w} \}$ accordingly.
GaussianAD passes information throughout the model with the sparse yet comprehensive 3D Gaussian representation, providing more knowledge to the decision-making process.
The overall framework of our GaussianAD is formulated as follows:
\begin{equation}\label{eqn: gaussianad}
\begin{aligned}
  & \{ \mathbf{o}^{T-H}, \cdots, \mathbf{o}^{T} \} \xrightarrow{} \mathbf{r}^T (\xrightarrow{} \mathbf{r}^T, \mathbf{d}^T) \xrightarrow{} \\
  &  \{ \mathbf{r}^T, \mathbf{r}^{T+1}, \cdots, \mathbf{r}^{T+F} \} \xrightarrow{} \{ \mathbf{w}^{T+1}, \cdots, \mathbf{w}^{T+F} \},
   \end{aligned}
\end{equation}
where $(\xrightarrow{} \mathbf{r}^T, \mathbf{d}^T)$ means that it is optional to incorporate additional perception supervision with $\mathbf{d}$ when available.

\begin{figure}[t]
\centering
\includegraphics[width=0.475\textwidth]{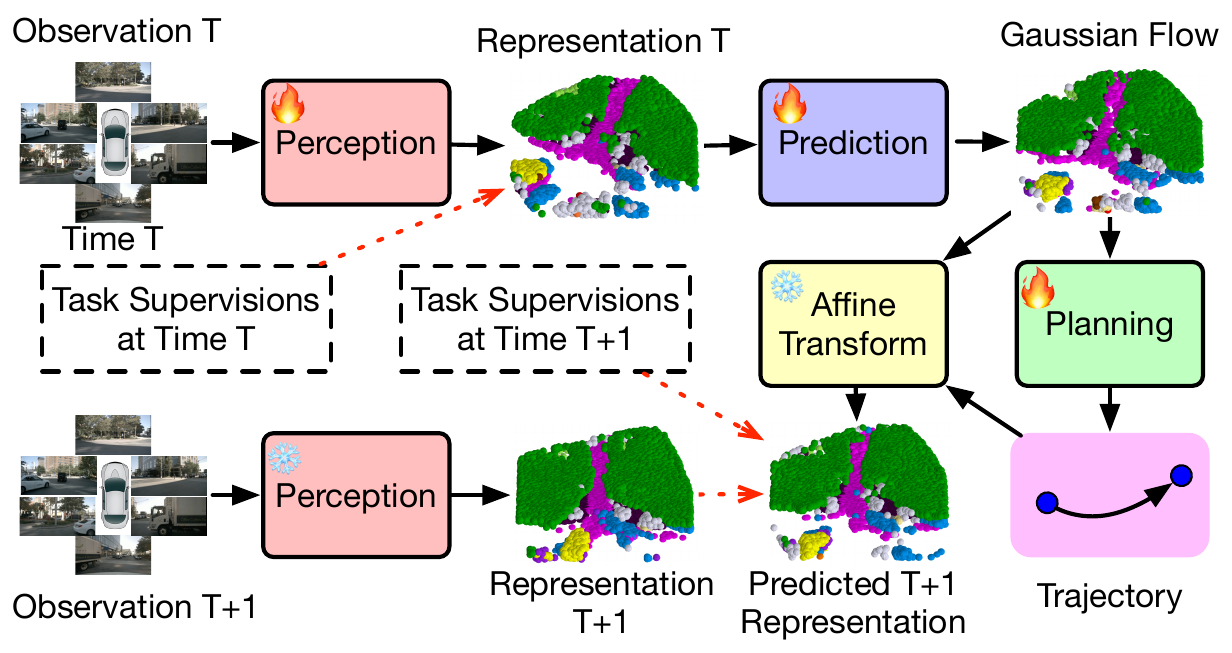}
\vspace{-8mm}
\caption{\textbf{Illustration of the training of our GaussianAD.}
Our framework can accommodate training data with different annotations by optionally imposing the corresponding supervisions on the scene representation.
Due to the explicit and structural nature of 3D Gaussians, we use global affine transformation to predict the future scene representations observed by the ego vehicle following the planned trajectory.
We can then use future perception labels or future scene representations obtained from future observations as the supervision.
They impose stronger constraints on the planned trajectory than the low-dimension trajectory discrepancy loss. 
}
\label{fig:training}
\vspace{-6mm}
\end{figure}

For training, we adaptively impose different perception losses on the scene descriptions $\mathbf{d}$ extracted from $\mathbf{r}$:
\begin{equation}\label{eqn: perc_loss}
\begin{aligned}
& J_{perc}(\mathbf{d}, \hat{\mathbf{d}}) = \lambda_{occ} J_{occ}(\mathbf{d}, \hat{\mathbf{d}}) + \lambda_{det} J_{det}(\mathbf{d}, \hat{\mathbf{d}}) \\
& + \lambda_{map} J_{map}(\mathbf{d}, \hat{\mathbf{d}}) + \lambda_{motion} J_{motion}(\mathbf{d}, \hat{\mathbf{d}}),
   \end{aligned}
\end{equation}
where $\lambda_{occ}$, $\lambda_{det}$, $\lambda_{map}$, and $\lambda_{motion}$ are balance factors and equal 0 if the supervision is not available.
$\hat{\mathbf{d}}$ denotes the ground-truth descriptions.
We use 3D occupancy prediction loss~\cite{gaussianformer} as $J_{occ}$, 3D detection loss~\cite{bevformer} as $J_{det}$, semantic map loss~\cite{beverse} as $J_{map}$, and motion loss~\cite{vad} as $J_{motion}$.

Due to the explicit representation of 3D Gaussians, we can use global affine transformation $t$ to simulate the scene representation $\widetilde{\mathbf{r}}$ observed at a certain given ego position $\mathbf{w}$. 
Having obtained the predicted future scene representations $\{ \mathbf{r}^T, \mathbf{r}^{T+1}, \cdots, \mathbf{r}^{T+F} \}$ with the proposed Gaussian flows, we simulate the future ego scene representations using the planned waypoints $\{ \mathbf{w}^{T+1}, \cdots, \mathbf{w}^{T+F} \}$:
\begin{equation}\label{eqn: affine}
\begin{aligned}
\{ \widetilde{\mathbf{r}} = t(\mathbf{r}, \mathbf{w}) \}^{F},
\end{aligned}
\end{equation}
where $^F$ denotes the future $F$ frames.
We then use the discrepancy between the simulated representations $\{\widetilde{\mathbf{r}}\}^F$ and ground truth representations $\{\hat{\mathbf{r}}\}^F$ as the loss:
\begin{equation}\label{eqn: pred_loss}
\begin{aligned}
& J_{pred}(\{\mathbf{r}\}^{F}, \{\hat{\mathbf{r}}\}^{F}, \{\hat{\mathbf{d}}\}^{F}) = \lambda_{re} J_{re}(\{\widetilde{\mathbf{r}}\}^{F}, \{\hat{\mathbf{r}}\}^{F}) \\
& + \lambda_{perc} J_{perc}(\{\widetilde{\mathbf{d}}(\widetilde{\mathbf{r}})\}^{F}, \{\hat{\mathbf{d}}\}^{F}), 
   \end{aligned}
\end{equation}
where $\lambda_{re}$ and $\lambda_{perc}$ are balance factors, and $J_{re}$ computes the discrepancy between two Gaussian representations.
$\{\hat{\mathbf{r}}\}^{F}$ can be computed from future observations $\{ \mathbf{o} \}$. 
$\widetilde{\mathbf{d}}(\widetilde{\mathbf{r}})$ denotes the predicted descpritions $\widetilde{\mathbf{d}}$ extracted from $\widetilde{\mathbf{r}}$.

The predicted future ego scene representations $\{ \widetilde{\mathbf{r}}\}^F$ also depend on the planned trajectories $\{ \mathbf{w}\}^F$ as in \eqref{eqn: affine}.
Therefore, we further adopt the prediction loss \eqref{eqn: pred_loss} for planning in addition to the conventional trajectory loss:
\begin{equation}\label{eqn: plan_loss}
\begin{aligned}
& J_{plan}(\{\mathbf{w}\}^{F}, \{\hat{\mathbf{w}}\}^{F}) = \lambda_{tra} J_{tra}(\{\mathbf{w}\}^{F}, \{\hat{\mathbf{w}}\}^{F}) \\
& + \lambda_{pred} J_{pred}(\{\mathbf{r}\}^{F}, \{\hat{\mathbf{r}}\}^{F}, \{\hat{\mathbf{d}}\}^{F}), 
   \end{aligned}
\end{equation}
where $\lambda_{tra}$ and $\lambda_{pred}$ are balance factors, and $\hat{\mathbf{w}}$ denotes the ground truth waypoint.
We adopt the trajectory losses from GenAD~\cite{genad} as $J_{tra}$.

The proposed GaussianAD is a flexible framework and can accommodate various cases with different available supervisions, as shown in Figure~\ref{fig:training}.
We train GaussianAD jointly with the following overall objective:
\begin{equation}\label{eqn: overall_losss}
\begin{aligned}
& J_{GaussianAD} = J_{perc} + J_{pred} + J_{plan},
\end{aligned}
\end{equation}
where $J_{perc}$, $J_{pred}$, and $J_{plan}$ can be customized for different scenarios.

For inference, GaussianAD accomplishes end-to-end driving using 3D Gaussian representation to efficiently pass information throughout the pipeline. 
It provides comprehensive knowledge for the decision-making process and maintains high efficiency with sparse computing.

\section{Experiments}

\begin{table*}[t]
\setlength{\tabcolsep}{0.010\linewidth}
\caption{\textbf{Open-looped motion planning results in comparison with state-of-the-art methods on the validation set of nuScenes~\cite{nuscenes}.} 
$\dagger$ denotes the results computed with an average of previous frames as adopted in VAD~\cite{vad}.
Aux. Sup. represents auxiliary supervision in addition to planning.
Avg. computes the average result of 1s, 2s, and 3s.
Bold numbers represent the best results.
}
\vspace{-3mm}
\centering
\begin{tabular}{l|cc|cccc|cccc}
\toprule
\multirow{2}{*}{Method} & \multirow{2}{*}{Input} & \multirow{2}{*}{Aux. Sup.} &
\multicolumn{4}{c|}{L2 (m) $\downarrow$} & 
\multicolumn{4}{c}{Collision Rate (\%) $\downarrow$} \\
&& & 1s & 2s & 3s & \cellcolor{gray!30}Avg. & 1s & 2s & 3s & \cellcolor{gray!30}Avg. \\
\midrule
IL~\cite{ratliff2006maximum} & LiDAR & None  & 0.44 & 1.15 & 2.47 & \cellcolor{gray!30}1.35 & 0.08 & 0.27 & 1.95 & \cellcolor{gray!30}0.77 \\
NMP~\cite{zeng2019nmp} & LiDAR & Box \& Motion & 0.53 & 1.25 & 2.67 & \cellcolor{gray!30}1.48 & 0.04 & 0.12 & 0.87 & \cellcolor{gray!30}0.34 \\
FF~\cite{hu2021ff} & LiDAR & Freespace  & 0.55 & 1.20 & 2.54 & \cellcolor{gray!30}1.43 & 0.06 & 0.17 & 1.07 & \cellcolor{gray!30}0.43 \\
EO~\cite{khurana2022eo} & LiDAR & Freespace  & 0.67 & 1.36 & 2.78 & \cellcolor{gray!30}1.60 & 0.04 & 0.09 & 0.88 & \cellcolor{gray!30}0.33 \\
\midrule
\midrule
ST-P3~\cite{hu2022stp3} & Camera & Map \& Box \& Depth & 1.33 & 2.11 & 2.90 & \cellcolor{gray!30}2.11 & 0.23 & 0.62 & 1.27 & \cellcolor{gray!30}0.71 \\
UniAD~\cite{uniad} & Camera & { \footnotesize Map \& Box \& Motion \& Tracklets \& Occ}  & {0.48} & 0.96 & 1.65 & \cellcolor{gray!30}1.03 & \underline{0.05} & \textbf{0.17} & \underline{0.71}& \cellcolor{gray!30}\textbf{0.31} \\
VAD-Tiny~\cite{vad}  & Camera & Map \& Box \& Motion  & 0.60 & 1.23 & 2.06 & \cellcolor{gray!30}1.30 & 0.31 & 0.53 & 1.33 & \cellcolor{gray!30}0.72 \\
VAD-Base~\cite{vad} & Camera & Map \& Box \& Motion & 0.54 & 1.15 & 1.98 & \cellcolor{gray!30}1.22 & \textbf{0.04} & 0.39 & 1.17 & \cellcolor{gray!30}0.53 \\
GenAD~\cite{genad} & Camera & Map \& Box \& Motion & \textbf{0.36} & 0.83& 1.55& \cellcolor{gray!30}0.91& 0.06 & \underline{0.23} & 1.00 & \cellcolor{gray!30}0.43  \\
\textbf{GaussianAD} & Camera & 3D-Occ \& Map \& Box \& Motion & \underline{0.40}& \textbf{0.64}& \textbf{0.88}& \cellcolor{gray!30}\textbf{0.64}& 0.09 & 0.38& 0.81& \cellcolor{gray!30}\underline{0.42}\\
\midrule
OccWorld~\cite{occworld} & Camera & 3D-Occ & 0.52 & 1.27 & 2.41 & \cellcolor{gray!30}1.40 & \textbf{0.12} & 0.40 & 2.08 & \cellcolor{gray!30}0.87  \\
OccNet~\cite{tong2023scene} &   Camera & 3D-Occ \& Map \& Box  & 1.29 & 2.13 & 2.99 & \cellcolor{gray!30}2.14 & 0.21 & 0.59 & 1.37 & \cellcolor{gray!30}0.72  \\
\textbf{GaussianAD} & Camera & 3D-Occ \& Map \& Box & \underline{0.40}& \underline{0.66}& \underline{0.92}& \cellcolor{gray!30}\underline{0.66}& 0.49& 0.38& \textbf{0.61}& \cellcolor{gray!30}0.49 \\
\midrule
\midrule
\color{gray}VAD-Tiny$^\dagger$~\cite{vad}  & \color{gray}Camera & \color{gray}Map \& Box \& Motion  & \color{gray}0.46 & \color{gray}0.76 & \color{gray}1.12 & \color{gray}\cellcolor{gray!30}0.78 & \color{gray}0.21 & \color{gray}0.35 & \color{gray}0.58 & \color{gray}\cellcolor{gray!30}0.38 \\
\color{gray}VAD-Base$^\dagger$~\cite{vad} & \color{gray}Camera & \color{gray}Map \& Box \& Motion & \color{gray}0.41 & \color{gray}0.70 & \color{gray}1.05 & \color{gray}\cellcolor{gray!30}0.72 & \color{gray}\textbf{0.07} & \color{gray}0.17 & \color{gray}\underline{0.41} & \color{gray}\cellcolor{gray!30}\underline{0.22} \\
\color{gray}{OccWorld-D}$^\dagger$~\cite{occworld} & \color{gray}Camera & \color{gray}3D-Occ & \color{gray}0.39 & \color{gray}0.73 & \color{gray}1.18 & \color{gray}\cellcolor{gray!30} 0.77 & \color{gray}0.11 & \color{gray}\underline{0.19} & \color{gray}0.67 & \color{gray}\cellcolor{gray!30} 0.32  \\
\color{gray}GenAD$^\dagger$~\cite{genad} & \color{gray}Camera & \color{gray}Map \& Box \& Motion & \color{gray}\textbf{0.28} & \color{gray}\underline{0.49}& \color{gray}\underline{0.78}& \color{gray}\cellcolor{gray!30}\underline{0.52}& \color{gray}\underline{0.08} & \color{gray}\textbf{0.14} & \color{gray}\textbf{0.34} & \color{gray}\cellcolor{gray!30}\textbf{0.19}  \\
\midrule
\color{gray}\textbf{GaussianAD}$^\dagger$ & \color{gray}Camera & \color{gray}3D-Occ \& \color{gray}Map \& \color{gray}Box   & \color{gray}\underline{0.34}& \color{gray}\textbf{0.47}& \color{gray}\textbf{0.60}& \cellcolor{gray!30}\color{gray}\textbf{0.47}& \color{gray}0.49& \color{gray}0.49& \color{gray}0.51& \cellcolor{gray!30}\color{gray}0.50\\
\bottomrule
\end{tabular}%
\label{tab:sota-plan}
\vspace{-7mm}
\end{table*}

\subsection{Datasets}

We conducted a series of experiments using the widely used nuScenes~\cite{nuscenes} dataset to evaluate our GaussianAD.
The nuScenes dataset consists of 1000 driving sequences, each providing 20 seconds of video captured by both RGB and LiDAR sensors. 
They provide data with a rate of 20Hz but only supply annotations for the keyframes at 2Hz, including labels for the semantic map construction and 3D object detection tasks.
The recent SurroundOcc~\cite{surroundocc} further complements nuScenes with 3D semantic occupancy annotations.
It assigns each voxel with a label of 18 categories including 16 semantic classes, 1 empty class, and 1 unknown class.

\subsection{Evaluation Metrics}
We evaluate the planning performance of our GaussianAD using the L2 displacement error and collision rate for fair comparisons with existing end-to-end methods~\cite{hu2022stp3, uniad, genad, occworld}. 
The L2 displacement error quantifies the difference between the planned and the ground-truth trajectory, computed as the L2 distance. 
The collision rate indicates the frequency with which the autonomous vehicle collides with other agents while following the planned path. 
For evaluation, we use a 2-second 5-frame history as input and compute the metric at future time steps of 1s, 2s, and 3s.

\begin{table}[t] \small
\setlength{\tabcolsep}{0.005\linewidth}
\caption{\textbf{Comparisons on 3D perception.} We report results on the 3D object detection, 3D occupancy prediction, and planning tasks.
$^\dagger$ denotes using a perception region of 51.2m $\times$ 51.2m.
}
\vspace{-3mm}
\begin{tabular}{l|c|cc|cc}
\toprule
\multirow{2}{*}{Method} &
\multicolumn{1}{c|}{Detection} & \multicolumn{2}{c|}{Occupancy} & \multicolumn{2}{c}{Planning} \\
 & mAP $\uparrow$ & mIoU $\uparrow$ & IoU $\uparrow$ & Avg. L2 $\downarrow$ & Avg. CR. $\downarrow$\\
\midrule
VAD~\cite{vad} & 0.27 & - & - & 1.30 & 0.72 \\
GenAD~\cite{genad} & \textbf{0.29} & - & - & 0.91 & 0.43\\
{TPVFormer$^\dagger$}~\cite{tpvformer} & - & 17.10 & 30.86 & - & -   \\
{\scriptsize GaussianFormer$^\dagger$}~\cite{gaussianformer} & - & 19.10 & 29.83 & - & -   \\
{SurroundOcc$^\dagger$}~\cite{surroundocc} & - & 20.30 & 31.49 & - & -   \\
\textbf{GaussianAD} & 0.19 & \textbf{22.12} & \textbf{33.81} & \textbf{0.64}& \textbf{0.42}\\
\bottomrule
\end{tabular}%
\label{tab:perception}
\vspace{-7mm}
\end{table}

\begin{table*}[t]
\setlength{\tabcolsep}{0.008\linewidth}
\caption{\textbf{Results of the scene prediction performance.}
We report results on the 4D occupancy forecasting task~\cite{occworld}.
Aux. Sup. represents auxiliary supervision in addition to planning.
Avg. computes the average result of 1s, 2s, and 3s.
$^*$ denotes using an end-to-end model.
}
\vspace{-3mm}
\centering
\begin{tabular}{l|cc|ccccc|ccccc}
\toprule
\multirow{2}{*}{Method} & \multirow{2}{*}{Input} & \multirow{2}{*}{Aux. Sup.} &
\multicolumn{5}{c|}{mIoU (\%) $\uparrow$} & 
\multicolumn{5}{c}{IoU (\%) $\uparrow$}
\\
&& & 0s & 1s & 2s & 3s & \cellcolor{gray!30}Avg. & 0s & 1s & 2s & 3s & \cellcolor{gray!30}Avg. \\
\midrule
{Copy\&Paste} & 3D-Occ & None & 66.38 & 14.91 & 10.54 & 8.52 & \cellcolor{gray!30}11.33 & 62.29 & 24.47 & 19.77 & 17.31 & \cellcolor{gray!30}20.52 \\
{OccWorld-O}~\cite{occworld} & 3D-Occ & None & \textbf{66.38} & \textbf{25.78} & \textbf{15.14} & \textbf{10.51} & \cellcolor{gray!30}\textbf{17.14}  & \textbf{62.29} & \textbf{34.63} & \textbf{25.07} & \textbf{20.18} & \cellcolor{gray!30}\textbf{26.63} \\
\hline
{OccWorld-T}~\cite{occworld} & Camera & Semantic LiDAR & 7.21 & 4.68 &  3.36 & 2.63 & \cellcolor{gray!30}3.56 & 10.66 & 9.32 & 8.23 & 7.47 & \cellcolor{gray!30}8.34 \\
{OccWorld-S}~\cite{occworld} & Camera & None & 0.27 & 0.28 & 0.26 & 0.24 & \cellcolor{gray!30}0.26 & 4.32 & 5.05 & 5.01 & 4.95 & \cellcolor{gray!30}5.00 \\
\hline
{OccWorld-D}~\cite{occworld} & Camera & 3D-Occ & 18.63 & 11.55 & 8.10 & 6.22 & \cellcolor{gray!30}8.62 & 22.88 & 18.90 & 16.26 & 14.43 & \cellcolor{gray!30}16.53 \\
\textbf{GaussianAD}$^*$ & Camera & 3D-Occ & 15.87& 6.29& 5.36& 4.58& \cellcolor{gray!30}5.41& 29.35& 14.13& 14.09& 14.04& \cellcolor{gray!30}14.30 \\
\bottomrule
\end{tabular}%
\label{forecast}
\vspace{-4mm}
\end{table*}

\begin{table*}[t]
\setlength{\tabcolsep}{0.014\linewidth}
\caption{\textbf{Effect of using different auxiliary supervision signals.} 
We analyze the effect of using additional 3D occupancy, 3D detection, map construction, motion prediction, and scene prediction labels as supervision.
Bold numbers represent the best results.
$^*$ represents that the use of the proposed flow-based prediction does not require additional annotations.
}
\vspace{-3mm}
\begin{tabular}{ccccc|cccc|cccc}
\toprule
\multirow{2}{*}{Occupancy} &  \multirow{2}{*}{Detection} & \multirow{2}{*}{Map} & \multirow{2}{*}{Motion}  & \multirow{2}{*}{Prediction$^*$} &
\multicolumn{4}{c|}{L2 (m) $\downarrow$} & 
\multicolumn{4}{c}{Collision Rate (\%) $\downarrow$}  \\
&&&& & 1s & 2s & 3s & \cellcolor{gray!30}Avg. & 1s & 2s & 3s & \cellcolor{gray!30}Avg.  \\
\midrule
$\times$ & $\checkmark$& $\checkmark$& $\times$ & $\times$ & 0.49& 0.75& 1.00& \cellcolor{gray!30}0.74& 0.28& 0.51& 1.12& \cellcolor{gray!30}0.63\\
$\checkmark$ & $\times$ & $\times$  & $\times$ & $\times$ & 0.41& 0.67& 0.91& \cellcolor{gray!30}0.66& 0.26& 0.43& 1.19& \cellcolor{gray!30}0.62\\
$\checkmark$ & $\checkmark$ & $\times$  & $\times$ & $\times$  & 0.41& 0.67& 0.92& \cellcolor{gray!30}0.66& 0.24& 0.68& 0.83& \cellcolor{gray!30}0.58\\
$\checkmark$ & $\checkmark$ & $\checkmark$ & $\times$  & $\times$ & 0.48& 0.67& 0.89& \cellcolor{gray!30}0.68& 0.16& 0.78& 0.81& \cellcolor{gray!30}0.58\\
$\checkmark$ & $\checkmark$ & $\checkmark$ & $\checkmark$  & $\times$ & \textbf{0.40}& \textbf{0.64}& \textbf{0.88}& \cellcolor{gray!30}\textbf{0.64}& \textbf{0.09}& \textbf{0.38}& 0.81& \cellcolor{gray!30}\textbf{0.42}\\
$\checkmark$ & $\checkmark$ & $\checkmark$ & $\times$& $\checkmark$ & \textbf{0.40}& 0.66& 0.92& \cellcolor{gray!30}0.66& 0.49& \textbf{0.38}& \textbf{0.61}& \cellcolor{gray!30}0.49\\
\bottomrule
\end{tabular}%
\label{tab:supervision}
\vspace{-7mm}
\end{table*}

\begin{table*}[t]
\vspace{-3mm}
\caption{\textbf{Effect of further Gaussian pruning.} 
We report results on the 3D occupancy and planning tasks.
}
\vspace{-3mm}
\centering
\begin{tabular}{c|cc|c|cccc|cccc}
\toprule
\multirow{2}{*}{\#Gaussians} & \multicolumn{2}{c|}{Occupancy} & \multicolumn{1}{c|}{Detection} & \multicolumn{4}{c|}{Planning L2 (m) $\downarrow$} & 
\multicolumn{4}{c}{Planning Collision Rate (\%) $\downarrow$}  \\
& mIoU$\uparrow$ & IoU$\uparrow$ & mAP$\uparrow$ & 1s & 2s & 3s & \cellcolor{gray!30}Avg. & 1s & 2s & 3s & \cellcolor{gray!30}Avg.  \\
\midrule
25600 & \textbf{21.22} & \textbf{33.81} & \textbf{0.16} & 0.48 & \textbf{0.67} & \textbf{0.89} & \cellcolor{gray!30}\textbf{0.68}& 0.16 & 0.78& 0.81& \cellcolor{gray!30}0.58 \\
20480 (-20\%) & 21.01& 33.61& \textbf{0.16} & 0.45& 0.69& 0.94& \cellcolor{gray!30}0.69& \textbf{0.08} & \textbf{0.29} & 1.29 & \cellcolor{gray!30}0.55\\
15360 (-40\%) & 20.57 & 33.77& 0.15& \textbf{0.43} & 0.68& 0.93& \cellcolor{gray!30}\textbf{0.68} & 0.28& 0.46& \textbf{0.59} & \cellcolor{gray!30}\textbf{0.44} \\
\bottomrule
\end{tabular}%
\label{tab:number}
\vspace{-4mm}
\end{table*}

\subsection{Implementation Details}
We employ ResNet101-DCN~\cite{resnet} with pre-trained weights from FCOS3D~\cite{fcos3d} as the backbone and additionaly use a feature pyramid network~\cite{FPN} to generate multi-scale image features. 
Our model takes as input images with a resolution of 1600 $\times$ 900 and sets the default number of Gaussians to 25600.
In the training stage, we take the AdamW~\cite{adamw} with a weight decay of 0.01 as the optimizer. The learning rate begins with 2e-4 and decreases according to a cosine schedule. By default, our models are trained on 32 A100 GPUs with a batch size of 8 for 20 epochs.

\subsection{Results and Analysis}

\textbf{End-to-End Planning Results.}
We provide comparisons with state-of-the-art end-to-end autonomous driving models in Table~\ref{tab:sota-plan}.
Bold numbers and underlined numbers denote the best and next-best results, respectively.
We also report the metrics used in VAD~\cite{vad}, which computes the average results of all the previous frames at each time stamp.

Note that different methods use different input modalities and auxiliary supervision signals that may influence the performance.
Generally, LiDAR provides additional depth information that is critical for planning, especially when measuring the collision rate.
However, the LiDAR point clouds, though accurate, are usually sparse and lack more fine-grained information, yielding inferior performance.
For auxiliary supervision, motions are usually considered the most effective labels as they provide ground truth for safety-critical future predictions. 
Still, motions are relatively expensive to annotate while 3D occupancy labels can be automatically annotated using multi-frame LiDAR and 3D bounding boxes~\cite{surroundocc}.
Though our GaussianAD can accommodate different supervision signals, we replace motion with 3D occupancy as the most practical setting.

Table~\ref{tab:sota-plan} shows that our method achieves the best performance on the L2 metric and competitive results on the collision rate metric.
In particular, GaussianAD outperforms OccNet~\cite{tong2023scene} using the same supervision signals (i.e., 3D occupancy, maps, and 3D bounding boxes) by a large margin.
Despite the lack of motion labels, our GaussianAD predicts Gaussian flows to simulate future scenes, enabling the exploitation of perception labels for the motion task. 
This forces the model to consider more about future interactions, resulting in the large improvements over OccNet~\cite{tong2023scene}.

\begin{figure*}[t]
\centering
\includegraphics[width=1\textwidth]{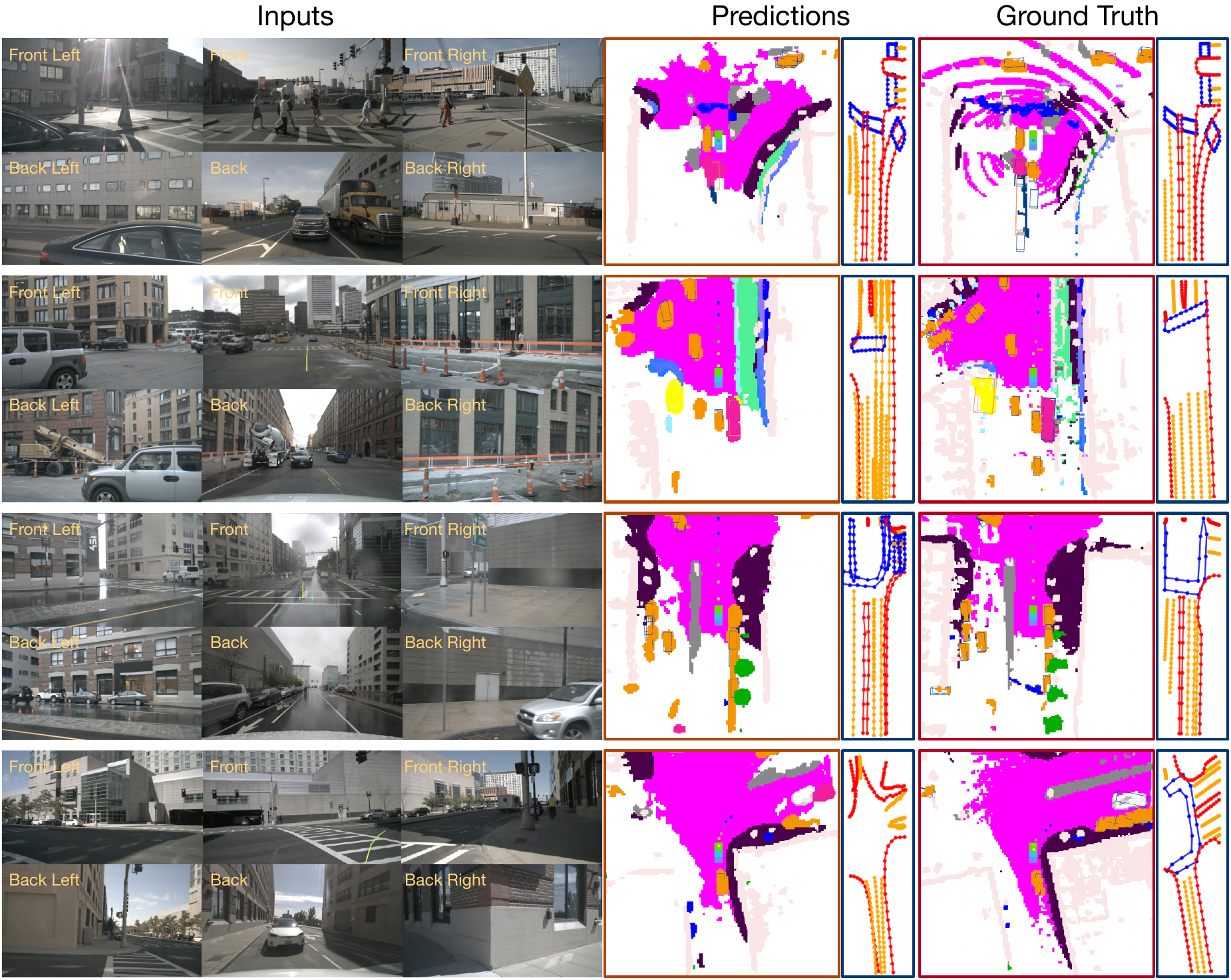}
\vspace{-8mm}
\caption{\textbf{Visualizations of the results of our GaussianAD.}
We include the 3D object detection and planning results in the 3D occupancy visualizations.
We also provide map visualizations.
(Better viewed on a monitor when zoomed in.)
}
\label{fig:vis}
\vspace{-7mm}
\end{figure*}

\textbf{3D Occupancy Prediction.}
We also provide results on other perception tasks though they are not the focus of this paper.
We adopt the mean average precision (mAP) for the 3D object detection task~\cite{bevformer, bevdepth}.
We use the mean intersection-over-union (mIoU) and intersection-over-union (IoU) for 3D occupancy prediction to measure the semantic and structural reconstruction quality, respectively.

Table~\ref{tab:perception} compares our GaussianAD with state-of-the-art end-to-end and 3D occupancy prediction methods.
GaussianAD shows good results on the 3D occupancy prediction task but underperforms existing end-to-end methods on 3D object detection.
This is because different perception tasks focus on different aspects of scene descriptions and could interfere with one another.
This explains the inferior performance of our method on the collision metric, which requires accurate perception of other agents to avoid collision.

\textbf{4D Occupancy Forecasting.}
By predicting a 3D flow for each Gaussian and performing the affine transformation using the planned trajectory, GaussianAD is able to forecast future scenes and perform perception on them.
We evaluate the prediction ability of GaussianAD on the 4D occupancy forecasting task~\cite{occworld} and measure the 3D occupancy quality (mIoU and IoU) at the future 1s, 2s, and 3s.

Table~\ref{forecast} shows that our GaussianAD can effectively predict forecast future 3D occupancy.
Note that our GaussianAD is an end-to-end model that performs multiple tasks simultaneously, while OccWorld~\cite{occworld} specifically targets this task.
Also, our forecasting does not consider the completion of newly observed areas (due to the ego car moving forward), leading to inferior performance.
GaussianAD still demonstrates non-trivial 4D forecasting results, verifying the effectiveness of the proposed Gaussian flows.

\textbf{Effect of Different Supervision Signals.}
As our model can adapt to different training signals for different tasks, we conducted an ablation study to analyze the effect of using different auxiliary supervision, as shown in Table~\ref{tab:supervision}.
We study the planning performance with a combination of 3D occupancy, 3D detection, map construction, motion prediction, and scene prediction supervision.
We see that our GaussianAD delivers consistent performance with different supervision combinations, and using more supervision signals generally improves the performance.
The use of motion supervision is particularly effective for the collision rate metric since it provides guidance on potential future overlap of trajectories.
Still, using the proposed flow-based scene prediction supervision achieves similar improvements,
which only requires future perception labels and introduces no additional annotations.

\textbf{3D Gaussian Pruning.}
We also analyze the effect of further pruning the Gaussians to reduce redundancy, as shown in Table~\ref{tab:number}.
We perform pruning by ordering the Gaussians according to their semantic confidence (i.e., the largest probability in the logits) and pruning the smallest ones.
We observe that Gaussian pruning slightly decreases the performance of perception tasks and yet improves the planning performance, demonstrating the potential of our framework.

\textbf{Visualizations.}
Figure~\ref{fig:vis} provides a visualization of the outputs of GaussianAD, which effectively perceives the surroundings and makes correct decisions in various scenarios.

\section{Conclusion}
We have presented a Gaussian-centric framework for vision-based end-to-end autonomous driving. 
To preserve more comprehensive information, we employ 3D Gaussians as the scene representation and adopt Gaussian flows to effectively predict future evolutions.
Our framework offers flexibility to accommodate different training data with various annotations.
We have conducted extensive experiments on the widely used nuScenes and demonstrated competitive performance on various tasks including ent-to-end planning and 4D occupancy forecasting.
It is interesting to explore larger-scale end-to-end models based on 3D Gaussian scene representation trained with more diverse data.

\textbf{Limitations.} GaussianAD cannot predict accurate scene evolutions since it does not consider newly observed areas.


\small
\begin{thebibliography}{66}
\providecommand{\natexlab}[1]{#1}
\providecommand{\url}[1]{\texttt{#1}}
\expandafter\ifx\csname urlstyle\endcsname\relax
  \providecommand{\doi}[1]{doi: #1}\else
  \providecommand{\doi}{doi: \begingroup \urlstyle{rm}\Url}\fi

\bibitem[Bouchard et~al.(2022)Bouchard, Sedwards, and Czarnecki]{bouchard2022rule}
Fr{\'e}d{\'e}ric Bouchard, Sean Sedwards, and Krzysztof Czarnecki.
\newblock A rule-based behaviour planner for autonomous driving.
\newblock In \emph{IJCRR}, pages 263--279, 2022.

\bibitem[Bronstein et~al.(2022)Bronstein, Palatucci, Notz, White, Kuefler, Lu, Paul, Nikdel, Mougin, Chen, et~al.]{bronstein2022hierarchical}
Eli Bronstein, Mark Palatucci, Dominik Notz, Brandyn White, Alex Kuefler, Yiren Lu, Supratik Paul, Payam Nikdel, Paul Mougin, Hongge Chen, et~al.
\newblock Hierarchical model-based imitation learning for planning in autonomous driving.
\newblock In \emph{IROS}, pages 8652--8659, 2022.

\bibitem[Caesar et~al.(2020)Caesar, Bankiti, Lang, Vora, Liong, Xu, Krishnan, Pan, Baldan, and Beijbom]{nuscenes}
Holger Caesar, Varun Bankiti, Alex~H Lang, Sourabh Vora, Venice~Erin Liong, Qiang Xu, Anush Krishnan, Yu Pan, Giancarlo Baldan, and Oscar Beijbom.
\newblock nuscenes: A multimodal dataset for autonomous driving.
\newblock In \emph{CVPR}, 2020.

\bibitem[Cao and de~Charette(2023)]{cao2023scenerf}
Anh-Quan Cao and Raoul de Charette.
\newblock Scenerf: Self-supervised monocular 3d scene reconstruction with radiance fields.
\newblock In \emph{ICCV}, pages 9387--9398, 2023.

\bibitem[Chai et~al.(2019)Chai, Sapp, Bansal, and Anguelov]{chai2019multipath}
Yuning Chai, Benjamin Sapp, Mayank Bansal, and Dragomir Anguelov.
\newblock Multipath: Multiple probabilistic anchor trajectory hypotheses for behavior prediction.
\newblock \emph{arXiv preprint arXiv:1910.05449}, 2019.

\bibitem[Chen et~al.(2020)Chen, Zhou, Koltun, and Kr{\"a}henb{\"u}hl]{chen2020lbc}
Dian Chen, Brady Zhou, Vladlen Koltun, and Philipp Kr{\"a}henb{\"u}hl.
\newblock Learning by cheating.
\newblock 2020.

\bibitem[Chen et~al.(2023)Chen, Liu, Zhang, Qi, and Jia]{voxelnext}
Yukang Chen, Jianhui Liu, Xiangyu Zhang, Xiaojuan Qi, and Jiaya Jia.
\newblock Voxelnext: Fully sparse voxelnet for 3d object detection and tracking.
\newblock \emph{arXiv preprint arXiv:2303.11301}, 2023.

\bibitem[Cheng et~al.(2022)Cheng, Xin, Wang, and Liu]{cheng2022mpnp}
Jie Cheng, Ren Xin, Sheng Wang, and Ming Liu.
\newblock Mpnp: Multi-policy neural planner for urban driving.
\newblock In \emph{IROS}, pages 10549--10554, 2022.

\bibitem[Cheng et~al.(2023)Cheng, Chen, Mei, Yang, Li, and Liu]{cheng2023rethinking}
Jie Cheng, Yingbing Chen, Xiaodong Mei, Bowen Yang, Bo Li, and Ming Liu.
\newblock Rethinking imitation-based planner for autonomous driving.
\newblock \emph{arXiv preprint arXiv:2309.10443}, 2023.

\bibitem[Codevilla et~al.(2019)Codevilla, Santana, L{\'o}pez, and Gaidon]{codevilla2019cilrs}
Felipe Codevilla, Eder Santana, Antonio~M L{\'o}pez, and Adrien Gaidon.
\newblock Exploring the limitations of behavior cloning for autonomous driving.
\newblock 2019.

\bibitem[Couto and Antonelo(2023)]{couto2023hierarchical}
Gustavo Claudio~Karl Couto and Eric~Aislan Antonelo.
\newblock Hierarchical generative adversarial imitation learning with mid-level input generation for autonomous driving on urban environments.
\newblock \emph{arXiv preprint arXiv:2302.04823}, 2023.

\bibitem[Dauner et~al.(2023)Dauner, Hallgarten, Geiger, and Chitta]{Dauner2023CORL}
Daniel Dauner, Marcel Hallgarten, Andreas Geiger, and Kashyap Chitta.
\newblock Parting with misconceptions about learning-based vehicle motion planning.
\newblock In \emph{CoRL}, 2023.

\bibitem[Gu et~al.(2022)Gu, Hu, Zhang, Chen, Wang, Wang, and Zhao]{gu2022vip3d}
Junru Gu, Chenxu Hu, Tianyuan Zhang, Xuanyao Chen, Yilun Wang, Yue Wang, and Hang Zhao.
\newblock Vip3d: End-to-end visual trajectory prediction via 3d agent queries.
\newblock \emph{arXiv preprint arXiv:2208.01582}, 2022.

\bibitem[Guo et~al.(2023)Guo, Jing, Chen, and Pan]{guo2023ccil}
Ke Guo, Wei Jing, Junbo Chen, and Jia Pan.
\newblock Ccil: Context-conditioned imitation learning for urban driving.
\newblock \emph{arXiv preprint arXiv:2305.02649}, 2023.

\bibitem[He et~al.(2016)He, Zhang, Ren, and Sun]{resnet}
Kaiming He, Xiangyu Zhang, Shaoqing Ren, and Jian Sun.
\newblock Deep residual learning for image recognition.
\newblock In \emph{CVPR}, 2016.

\bibitem[Hu et~al.(2021{\natexlab{a}})Hu, Murez, Mohan, Dudas, Hawke, Badrinarayanan, Cipolla, and Kendall]{hu2021fiery}
Anthony Hu, Zak Murez, Nikhil Mohan, Sof{\'\i}a Dudas, Jeffrey Hawke, Vijay Badrinarayanan, Roberto Cipolla, and Alex Kendall.
\newblock Fiery: Future instance prediction in bird's-eye view from surround monocular cameras.
\newblock In \emph{ICCV}, 2021{\natexlab{a}}.

\bibitem[Hu et~al.(2021{\natexlab{b}})Hu, Huang, Dolan, Held, and Ramanan]{hu2021ff}
Peiyun Hu, Aaron Huang, John Dolan, David Held, and Deva Ramanan.
\newblock Safe local motion planning with self-supervised freespace forecasting.
\newblock In \emph{CVPR}, 2021{\natexlab{b}}.

\bibitem[Hu et~al.(2022)Hu, Chen, Wu, Li, Yan, and Tao]{hu2022stp3}
Shengchao Hu, Li Chen, Penghao Wu, Hongyang Li, Junchi Yan, and Dacheng Tao.
\newblock St-p3: End-to-end vision-based autonomous driving via spatial-temporal feature learning.
\newblock In \emph{ECCV}, 2022.

\bibitem[Hu et~al.(2023)Hu, Yang, Chen, Li, Sima, Zhu, Chai, Du, Lin, Wang, et~al.]{uniad}
Yihan Hu, Jiazhi Yang, Li Chen, Keyu Li, Chonghao Sima, Xizhou Zhu, Siqi Chai, Senyao Du, Tianwei Lin, Wenhai Wang, et~al.
\newblock Planning-oriented autonomous driving.
\newblock In \emph{CVPR}, pages 17853--17862, 2023.

\bibitem[Huang et~al.(2021)Huang, Huang, Zhu, and Du]{bevdet}
Junjie Huang, Guan Huang, Zheng Zhu, and Dalong Du.
\newblock Bevdet: High-performance multi-camera 3d object detection in bird-eye-view.
\newblock \emph{arXiv preprint arXiv:2112.11790}, 2021.

\bibitem[Huang et~al.(2023{\natexlab{a}})Huang, Zheng, Zhang, Zhou, and Lu]{tpvformer}
Yuanhui Huang, Wenzhao Zheng, Yunpeng Zhang, Jie Zhou, and Jiwen Lu.
\newblock Tri-perspective view for vision-based 3d semantic occupancy prediction.
\newblock In \emph{CVPR}, pages 9223--9232, 2023{\natexlab{a}}.

\bibitem[Huang et~al.(2024{\natexlab{a}})Huang, Thammatadatrakoon, Zheng, Zhang, Du, and Lu]{gaussianformer-2}
Yuanhui Huang, Amonnut Thammatadatrakoon, Wenzhao Zheng, Yunpeng Zhang, Dalong Du, and Jiwen Lu.
\newblock Gaussianformer-2: Probabilistic gaussian superposition for efficient 3d occupancy prediction.
\newblock \emph{arXiv preprint arXiv:2412.04384}, 2024{\natexlab{a}}.

\bibitem[Huang et~al.(2024{\natexlab{b}})Huang, Zheng, Zhang, Zhou, and Lu]{gaussianformer}
Yuanhui Huang, Wenzhao Zheng, Yunpeng Zhang, Jie Zhou, and Jiwen Lu.
\newblock Gaussianformer: Scene as gaussians for vision-based 3d semantic occupancy prediction.
\newblock \emph{arXiv preprint arXiv:2405.17429}, 2024{\natexlab{b}}.

\bibitem[Huang et~al.(2022)Huang, Wu, and Lv]{huang2022efficient}
Zhiyu Huang, Jingda Wu, and Chen Lv.
\newblock Efficient deep reinforcement learning with imitative expert priors for autonomous driving.
\newblock \emph{TNNLS}, 2022.

\bibitem[Huang et~al.(2023{\natexlab{b}})Huang, Liu, and Lv]{huang2023gameformer}
Zhiyu Huang, Haochen Liu, and Chen Lv.
\newblock Gameformer: Game-theoretic modeling and learning of transformer-based interactive prediction and planning for autonomous driving.
\newblock \emph{arXiv preprint arXiv:2303.05760}, 2023{\natexlab{b}}.

\bibitem[Huang et~al.(2023{\natexlab{c}})Huang, Liu, Wu, and Lv]{huang2023dipp}
Zhiyu Huang, Haochen Liu, Jingda Wu, and Chen Lv.
\newblock Differentiable integrated motion prediction and planning with learnable cost function for autonomous driving.
\newblock \emph{TNNLS}, 2023{\natexlab{c}}.

\bibitem[Jiang et~al.(2022)Jiang, Chen, Wang, Liao, Cheng, Chen, Zhou, Zhang, Liu, and Huang]{jiang2022pip}
Bo Jiang, Shaoyu Chen, Xinggang Wang, Bencheng Liao, Tianheng Cheng, Jiajie Chen, Helong Zhou, Qian Zhang, Wenyu Liu, and Chang Huang.
\newblock Perceive, interact, predict: Learning dynamic and static clues for end-to-end motion prediction.
\newblock \emph{arXiv preprint arXiv:2212.02181}, 2022.

\bibitem[Jiang et~al.(2023)Jiang, Chen, Xu, Liao, Chen, Zhou, Zhang, Liu, Huang, and Wang]{vad}
Bo Jiang, Shaoyu Chen, Qing Xu, Bencheng Liao, Jiajie Chen, Helong Zhou, Qian Zhang, Wenyu Liu, Chang Huang, and Xinggang Wang.
\newblock Vad: Vectorized scene representation for efficient autonomous driving.
\newblock \emph{arXiv preprint arXiv:2303.12077}, 2023.

\bibitem[Khurana et~al.(2022)Khurana, Hu, Dave, Ziglar, Held, and Ramanan]{khurana2022eo}
Tarasha Khurana, Peiyun Hu, Achal Dave, Jason Ziglar, David Held, and Deva Ramanan.
\newblock Differentiable raycasting for self-supervised occupancy forecasting.
\newblock In \emph{ECCV}, 2022.

\bibitem[Li et~al.(2022{\natexlab{a}})Li, Wang, Wang, and Zhao]{li2022hdmapnet}
Qi Li, Yue Wang, Yilun Wang, and Hang Zhao.
\newblock Hdmapnet: An online hd map construction and evaluation framework.
\newblock In \emph{ICRA}, 2022{\natexlab{a}}.

\bibitem[Li et~al.(2022{\natexlab{b}})Li, Ge, Yu, Yang, Wang, Shi, Sun, and Li]{bevdepth}
Yinhao Li, Zheng Ge, Guanyi Yu, Jinrong Yang, Zengran Wang, Yukang Shi, Jianjian Sun, and Zeming Li.
\newblock Bevdepth: Acquisition of reliable depth for multi-view 3d object detection.
\newblock \emph{arXiv preprint arXiv:2206.10092}, 2022{\natexlab{b}}.

\bibitem[Li et~al.(2022{\natexlab{c}})Li, Wang, Li, Xie, Sima, Lu, Yu, and Dai]{bevformer}
Zhiqi Li, Wenhai Wang, Hongyang Li, Enze Xie, Chonghao Sima, Tong Lu, Qiao Yu, and Jifeng Dai.
\newblock Bevformer: Learning bird's-eye-view representation from multi-camera images via spatiotemporal transformers.
\newblock In \emph{ECCV}, 2022{\natexlab{c}}.

\bibitem[Liang et~al.(2020)Liang, Yang, Hu, Chen, Liao, Feng, and Urtasun]{liang2020lanegcn}
Ming Liang, Bin Yang, Rui Hu, Yun Chen, Renjie Liao, Song Feng, and Raquel Urtasun.
\newblock Learning lane graph representations for motion forecasting.
\newblock In \emph{ECCV}, 2020.

\bibitem[Liao et~al.(2022)Liao, Chen, Wang, Cheng, Zhang, Liu, and Huang]{liao2022maptr}
Bencheng Liao, Shaoyu Chen, Xinggang Wang, Tianheng Cheng, Qian Zhang, Wenyu Liu, and Chang Huang.
\newblock Maptr: Structured modeling and learning for online vectorized hd map construction.
\newblock \emph{arXiv preprint arXiv:2208.14437}, 2022.

\bibitem[Lin et~al.(2017)Lin, Dollár, Girshick, He, Hariharan, and Belongie]{FPN}
Tsung-Yi Lin, Piotr Dollár, Ross Girshick, Kaiming He, Bharath Hariharan, and Serge Belongie.
\newblock Feature pyramid networks for object detection.
\newblock In \emph{CVPR}, 2017.

\bibitem[Liu et~al.(2022{\natexlab{a}})Liu, Huang, Wu, and Lv]{liu2022improved}
Haochen Liu, Zhiyu Huang, Jingda Wu, and Chen Lv.
\newblock Improved deep reinforcement learning with expert demonstrations for urban autonomous driving.
\newblock In \emph{TIV}, pages 921--928, 2022{\natexlab{a}}.

\bibitem[Liu et~al.(2021)Liu, Zhang, Fang, Jiang, and Zhou]{liu2021mmtrans}
Yicheng Liu, Jinghuai Zhang, Liangji Fang, Qinhong Jiang, and Bolei Zhou.
\newblock Multimodal motion prediction with stacked transformers.
\newblock In \emph{CVPR}, 2021.

\bibitem[Liu et~al.(2022{\natexlab{b}})Liu, Wang, Wang, and Zhao]{liu2022vectormapnet}
Yicheng Liu, Yue Wang, Yilun Wang, and Hang Zhao.
\newblock Vectormapnet: End-to-end vectorized hd map learning.
\newblock \emph{arXiv preprint arXiv:2206.08920}, 2022{\natexlab{b}}.

\bibitem[Loshchilov and Hutter(2019)]{adamw}
Ilya Loshchilov and Frank Hutter.
\newblock Decoupled weight decay regularization.
\newblock In \emph{ICLR}, 2019.

\bibitem[Ngiam et~al.(2021)Ngiam, Caine, Vasudevan, Zhang, Chiang, Ling, Roelofs, Bewley, Liu, Venugopal, et~al.]{ngiam2021scenetrans}
Jiquan Ngiam, Benjamin Caine, Vijay Vasudevan, Zhengdong Zhang, Hao-Tien~Lewis Chiang, Jeffrey Ling, Rebecca Roelofs, Alex Bewley, Chenxi Liu, Ashish Venugopal, et~al.
\newblock Scene transformer: A unified architecture for predicting multiple agent trajectories.
\newblock \emph{arXiv preprint arXiv:2106.08417}, 2021.

\bibitem[Phan-Minh et~al.(2020)Phan-Minh, Grigore, Boulton, Beijbom, and Wolff]{phan2020covernet}
Tung Phan-Minh, Elena~Corina Grigore, Freddy~A Boulton, Oscar Beijbom, and Eric~M Wolff.
\newblock Covernet: Multimodal behavior prediction using trajectory sets.
\newblock In \emph{CVPR}, 2020.

\bibitem[Philion and Fidler(2020)]{lss}
Jonah Philion and Sanja Fidler.
\newblock Lift, splat, shoot: Encoding images from arbitrary camera rigs by implicitly unprojecting to 3d.
\newblock In \emph{ECCV}, pages 194--210, 2020.

\bibitem[Pini et~al.(2023)Pini, Perone, Ahuja, Ferreira, Niendorf, and Zagoruyko]{pini2023safepathnet}
Stefano Pini, Christian~S Perone, Aayush Ahuja, Ana Sofia~Rufino Ferreira, Moritz Niendorf, and Sergey Zagoruyko.
\newblock Safe real-world autonomous driving by learning to predict and plan with a mixture of experts.
\newblock In \emph{ICRA}, pages 10069--10075, 2023.

\bibitem[Ratliff et~al.(2006)Ratliff, Bagnell, and Zinkevich]{ratliff2006maximum}
Nathan~D Ratliff, J~Andrew Bagnell, and Martin~A Zinkevich.
\newblock Maximum margin planning.
\newblock In \emph{ICML}, pages 729--736, 2006.

\bibitem[Reading et~al.(2021)Reading, Harakeh, Chae, and Waslander]{caddn}
Cody Reading, Ali Harakeh, Julia Chae, and Steven~L Waslander.
\newblock Categorical depth distribution network for monocular 3d object detection.
\newblock In \emph{CVPR}, 2021.

\bibitem[Scheel et~al.(2022)Scheel, Bergamini, Wolczyk, Osi{\'n}ski, and Ondruska]{scheel2022urban}
Oliver Scheel, Luca Bergamini, Maciej Wolczyk, B{\l}a{\.z}ej Osi{\'n}ski, and Peter Ondruska.
\newblock Urban driver: Learning to drive from real-world demonstrations using policy gradients.
\newblock In \emph{CoRL}, pages 718--728. PMLR, 2022.

\bibitem[Silva et~al.(2024)Silva, Bhashitha~Wannigama, Ragel, and Jayatilaka]{s2tpvformer}
Sathira Silva, Savindu Bhashitha~Wannigama, Roshan Ragel, and Gihan Jayatilaka.
\newblock S2tpvformer: Spatio-temporal tri-perspective view for temporally coherent 3d semantic occupancy prediction.
\newblock \emph{arXiv e-prints}, pages arXiv--2401, 2024.

\bibitem[Sun et~al.(2024)Sun, Lin, Shi, Zhang, Wu, and Zheng]{sparsedrive}
Wenchao Sun, Xuewu Lin, Yining Shi, Chuang Zhang, Haoran Wu, and Sifa Zheng.
\newblock Sparsedrive: End-to-end autonomous driving via sparse scene representation.
\newblock \emph{arXiv preprint arXiv:2405.19620}, 2024.

\bibitem[Tian et~al.(2023)Tian, Jiang, Yun, Wang, Wang, and Zhao]{tian2023occ3d}
Xiaoyu Tian, Tao Jiang, Longfei Yun, Yue Wang, Yilun Wang, and Hang Zhao.
\newblock Occ3d: A large-scale 3d occupancy prediction benchmark for autonomous driving.
\newblock \emph{arXiv preprint arXiv:2304.14365}, 2023.

\bibitem[Tong et~al.(2023)Tong, Sima, Wang, Chen, Wu, Deng, Gu, Lu, Luo, Lin, et~al.]{tong2023scene}
Wenwen Tong, Chonghao Sima, Tai Wang, Li Chen, Silei Wu, Hanming Deng, Yi Gu, Lewei Lu, Ping Luo, Dahua Lin, et~al.
\newblock Scene as occupancy.
\newblock In \emph{ICCV}, pages 8406--8415, 2023.

\bibitem[Treiber et~al.(2000)Treiber, Hennecke, and Helbing]{treiber2000idm}
Martin Treiber, Ansgar Hennecke, and Dirk Helbing.
\newblock Congested traffic states in empirical observations and microscopic simulations.
\newblock \emph{Physical review E}, 62\penalty0 (2):\penalty0 1805, 2000.

\bibitem[Vaswani et~al.(2017)Vaswani, Shazeer, Parmar, Uszkoreit, Jones, Gomez, Kaiser, and Polosukhin]{vaswani2017attention}
Ashish Vaswani, Noam Shazeer, Niki Parmar, Jakob Uszkoreit, Llion Jones, Aidan~N Gomez, {\L}ukasz Kaiser, and Illia Polosukhin.
\newblock Attention is all you need.
\newblock \emph{NeurIPS}, 2017.

\bibitem[Vitelli et~al.(2022)Vitelli, Chang, Ye, Ferreira, Wo{\l}czyk, Osi{\'n}ski, Niendorf, Grimmett, Huang, Jain, et~al.]{vitelli2022safetynet}
Matt Vitelli, Yan Chang, Yawei Ye, Ana Ferreira, Maciej Wo{\l}czyk, B{\l}a{\.z}ej Osi{\'n}ski, Moritz Niendorf, Hugo Grimmett, Qiangui Huang, Ashesh Jain, et~al.
\newblock Safetynet: Safe planning for real-world self-driving vehicles using machine-learned policies.
\newblock In \emph{ICRA}, pages 897--904, 2022.

\bibitem[Wang et~al.(2021)Wang, Zhu, Pang, and Lin]{fcos3d}
Tai Wang, Xinge Zhu, Jiangmiao Pang, and Dahua Lin.
\newblock Fcos3d: Fully convolutional one-stage monocular 3d object detection.
\newblock In \emph{ICCV}, 2021.

\bibitem[Wang et~al.(2023)Wang, Zhu, Xu, Zhang, Wei, Chi, Ye, Du, Lu, and Wang]{openoccupancy}
Xiaofeng Wang, Zheng Zhu, Wenbo Xu, Yunpeng Zhang, Yi Wei, Xu Chi, Yun Ye, Dalong Du, Jiwen Lu, and Xingang Wang.
\newblock Openoccupancy: A large scale benchmark for surrounding semantic occupancy perception.
\newblock \emph{arXiv preprint arXiv:2303.03991}, 2023.

\bibitem[Wei et~al.(2023)Wei, Zhao, Zheng, Zhu, Zhou, and Lu]{surroundocc}
Yi Wei, Linqing Zhao, Wenzhao Zheng, Zheng Zhu, Jie Zhou, and Jiwen Lu.
\newblock Surroundocc: Multi-camera 3d occupancy prediction for autonomous driving.
\newblock In \emph{ICCV}, pages 21729--21740, 2023.

\bibitem[Wen et~al.(2020)Wen, Lin, Darrell, Jayaraman, and Gao]{wen2020fighting}
Chuan Wen, Jierui Lin, Trevor Darrell, Dinesh Jayaraman, and Yang Gao.
\newblock Fighting copycat agents in behavioral cloning from observation histories.
\newblock \emph{NeurIPS}, 33:\penalty0 2564--2575, 2020.

\bibitem[Xie et~al.(2024)Xie, Zuo, Zheng, Zhang, Du, Zhou, Lu, and Zhang]{gpd-1}
Zixun Xie, Sicheng Zuo, Wenzhao Zheng, Yunpeng Zhang, Dalong Du, Jie Zhou, Jiwen Lu, and Shanghang Zhang.
\newblock Gpd-1: Generative pre-training for driving.
\newblock \emph{arXiv preprint arXiv:2412.08643}, 2024.

\bibitem[Ye et~al.(2023)Ye, Jing, Hu, Huang, Gao, Li, Wang, Guo, Xiao, Mao, et~al.]{ye2023fusionad}
Tengju Ye, Wei Jing, Chunyong Hu, Shikun Huang, Lingping Gao, Fangzhen Li, Jingke Wang, Ke Guo, Wencong Xiao, Weibo Mao, et~al.
\newblock Fusionad: Multi-modality fusion for prediction and planning tasks of autonomous driving.
\newblock \emph{arXiv preprint arXiv:2308.01006}, 2023.

\bibitem[Zeng et~al.(2019)Zeng, Luo, Suo, Sadat, Yang, Casas, and Urtasun]{zeng2019nmp}
Wenyuan Zeng, Wenjie Luo, Simon Suo, Abbas Sadat, Bin Yang, Sergio Casas, and Raquel Urtasun.
\newblock End-to-end interpretable neural motion planner.
\newblock In \emph{CVPR}, 2019.

\bibitem[Zhang et~al.(2024)Zhang, Wang, Zhu, Zhao, Chen, Zhang, Gong, Zhou, Zhang, Wang, et~al.]{sparsead}
Diankun Zhang, Guoan Wang, Runwen Zhu, Jianbo Zhao, Xiwu Chen, Siyu Zhang, Jiahao Gong, Qibin Zhou, Wenyuan Zhang, Ningzi Wang, et~al.
\newblock Sparsead: Sparse query-centric paradigm for efficient end-to-end autonomous driving.
\newblock \emph{arXiv preprint arXiv:2404.06892}, 2024.

\bibitem[Zhang et~al.(2022)Zhang, Zhu, Zheng, Huang, Huang, Zhou, and Lu]{beverse}
Yunpeng Zhang, Zheng Zhu, Wenzhao Zheng, Junjie Huang, Guan Huang, Jie Zhou, and Jiwen Lu.
\newblock Beverse: Unified perception and prediction in birds-eye-view for vision-centric autonomous driving.
\newblock \emph{arXiv preprint arXiv:2205.09743}, 2022.

\bibitem[Zheng et~al.(2024{\natexlab{a}})Zheng, Chen, Huang, Zhang, Duan, and Lu]{occworld}
Wenzhao Zheng, Weiliang Chen, Yuanhui Huang, Borui Zhang, Yueqi Duan, and Jiwen Lu.
\newblock Occworld: Learning a 3d occupancy world model for autonomous driving.
\newblock In \emph{ECCV}, 2024{\natexlab{a}}.

\bibitem[Zheng et~al.(2024{\natexlab{b}})Zheng, Song, Guo, and Chen]{genad}
Wenzhao Zheng, Ruiqi Song, Xianda Guo, and Long Chen.
\newblock Genad: Generative end-to-end autonomous driving.
\newblock In \emph{ECCV}, 2024{\natexlab{b}}.

\bibitem[Zhou et~al.(2021)Zhou, Wang, Liu, Jiang, Jiang, Tao, Miao, and Song]{zhou2021exploring}
Jinyun Zhou, Rui Wang, Xu Liu, Yifei Jiang, Shu Jiang, Jiaming Tao, Jinghao Miao, and Shiyu Song.
\newblock Exploring imitation learning for autonomous driving with feedback synthesizer and differentiable rasterization.
\newblock In \emph{IROS}, pages 1450--1457, 2021.

\bibitem[Zuo et~al.(2023)Zuo, Zheng, Huang, Zhou, and Lu]{pointocc}
Sicheng Zuo, Wenzhao Zheng, Yuanhui Huang, Jie Zhou, and Jiwen Lu.
\newblock Pointocc: Cylindrical tri-perspective view for point-based 3d semantic occupancy prediction.
\newblock \emph{arXiv preprint arXiv:2308.16896}, 2023.

\end{thebibliography}
\end{document}